\documentclass[journal]{IEEEtran}
 \pdfoutput=1
\usepackage{cite}
\usepackage{amsmath,amssymb,amsfonts}
\usepackage{algorithmic}
\usepackage{graphicx}
\usepackage{textcomp}
\usepackage{framed,multirow}
\usepackage{latexsym}
\usepackage{amsmath}
\usepackage{threeparttable}
\usepackage{float}
\usepackage{array}
\usepackage{hhline}
\usepackage{colortbl}
\usepackage{stfloats}
\usepackage{booktabs}
\usepackage{multirow}
\usepackage{enumerate}
\usepackage{diagbox}
\usepackage{url}
\usepackage{hyperref}
\usepackage{makecell,rotating}
\usepackage{caption}
\usepackage{color}
\usepackage{arydshln}
\usepackage[linesnumbered, ruled]{algorithm2e}
\DeclareUnicodeCharacter{2061}{}

\def\etal{\textit{et~al}.}

\DeclareGraphicsExtensions{.pdf,.jpeg,.png,.jpg}

\graphicspath{{fig/}}

\definecolor{mygray}{gray}{.9}

\definecolor{newcolor}{rgb}{.8,.349,.1}

\makeatletter

\newcommand{\Rmnum}[1]{\expandafter\@slowromancap\romannumeral #1@}
\makeatother

\title{Rethinking Mitosis Detection: Towards Diverse Data and Feature Representation}

\author{Hao Wang, Jiatai Lin, Danyi Li, Jing Wang, Bingchao Zhao, Zhenwei Shi, Xipeng Pan, Huadeng Wang, Bingbing Li, Changhong Liang, Guoqiang Han, Li Liang, Chu Han~\IEEEmembership{Member, IEEE}, Zaiyi Liu
\thanks{Hao Wang, Jiatai Lin and Guoqiang Han are with the School of Computer Science and Engineering, South China University of Technology, Guangzhou 510006, China.}
\thanks{Bingchao Zhao, Changhong Liang, Zaiyi Liu and Chu Han are with Department of Radiology, Guangdong Provincial People's Hospital (Guangdong Academy of Medical Sciences),Southern Medical University, Guangzhou 510080, China. Guangdong Provincial Key Laboratory of Artificial Intelligence in Medical Image Analysis and Application, Guangzhou 510080, China.}
\thanks{Danyi Li, Jing Wang and Li Liang are with the Department of Pathology, Nanfang Hospital and Basic Medical College, Southern Medical University, Guangzhou, Guangdong, China. Guangdong Province Key Laboratory of Molecular Tumor Pathology, Guangzhou, Guangdong, China.}
\thanks{Bingbing Li is with with the Department of Pathology, Guangdong Provincial People's Hospital Ganzhou Hospital, Ganzhou, China.}
\thanks{Xipeng Pan and Huadeng Wang are with the School of Computer Science and Information Security, Guilin University of Electronic Technology, Guilin 541004, China.}
\thanks{Corresponding author: Guoqiang Han, Li Liang, Chu Han, Zaiyi Liu.}
\thanks{The first three authors contributed equally.}}

\begin{document}
\maketitle

\IEEEtitleabstractindextext{\begin{abstract}
Mitosis detection is one of the fundamental tasks in computational pathology, which is extremely challenging due to the heterogeneity of mitotic cell. Most of the current studies solve the heterogeneity in the technical aspect by increasing the model complexity. However, lacking consideration of the biological knowledge and the complex model design may lead to the overfitting problem while limited the generalizability of the detection model. In this paper, we systematically study the morphological appearances in different mitotic phases as well as the ambiguous non-mitotic cells and identify that balancing the data and feature diversity can achieve better generalizability. Based on this observation, we propose a novel generalizable framework (MitDet) for mitosis detection. The data diversity is considered by the proposed diversity-guided sample balancing (DGSB). And the feature diversity is preserved by inter- and intra- class feature diversity-preserved module (InCDP). Stain enhancement (SE) module is introduced to enhance the domain-relevant diversity of both data and features simultaneously. Extensive experiments have demonstrated that our proposed model outperforms all the SOTA approaches in several popular mitosis detection datasets in both internal and external test sets using minimal annotation efforts with point annotations only. Comprehensive ablation studies have also proven the effectiveness of the rethinking of data and feature diversity balancing. By analyzing the results quantitatively and qualitatively, we believe that our proposed model not only achieves SOTA performance but also might inspire the future studies in new perspectives. Source code is at \url{https://github.com/Onehour0108/MitDet}.
\end{abstract}
}

\maketitle
\IEEEdisplaynontitleabstractindextext

\IEEEpeerreviewmaketitle

\section{Introduction}
Mitosis can reflect cell proliferation which is a crucial indicator of cancer grading and prognosis~\cite{2010Pathological}. Typically, pathologists count the number of mitotic cells from a fixed number of high power fields (HPFs) of the microscope~\cite{tellez2018whole}. However, manual assessment is time-consuming and subjective, highlighting the need for the automatic detection method to provide fast, accurate, and reproducible results for the pathologists in routine diagnosis. 

However, mitosis detection is an extremely challenging task for the following reasons. First, mitotic cells only account for an extremely small proportion of all the cells (around 0.3\% in MIDOG2021~\cite{aubreville2023mitosis} dataset), leading to dramatically imbalanced positive and negative samples. Second, differentiating mitotic and non-mitotic cells is difficult even for pathologists. Because non-mitotic cells are highly heterogeneous in malignant tumors and the morphological appearances of mitotic cells are also diverse. For example, there are four distinct phases of mitosis, including prophase, metaphase, anaphase, and telophase~\cite{2008The,Chao2019Weakly}. Third, domain shift can be usually observed across different medical institutions due to the inconsistent image acquisition protocols, especially the color variance~\cite{li2022domain}. Therefore, it is urge to propose a generalizable model for automatic mitosis detection.

Thanks to the rapid advancements of deep learning techniques, we have witnessed a great progress on mitosis detection~\cite{pan2021mitosis}. Until now, the existing models can be roughly divided into one-stage ones and multi-stages ones~\cite{aubreville2023mitosis}. One-stage models~\cite{Chao2019Weakly,han2021contextual} typically treat the mitotic cell detection task as a semantic segmentation task or an object detection task in an end-to-end manner. Multi-stage approaches~\cite{sohail2021mitotic,mahmood2020artificial,ccayir2022mitnet} first identify the candidates of mitotic cells, and then use a more sophisticated classifier to classify the candidate patches. However, restricted by the small size of the previous released datasets, the generalizability of the above-mentioned approaches cannot be well estimated. 
MIDOG2021 and MIDOG2022 challenges introduce two large-scale mitosis detection datasets with high sample diversity and domain discrepancy, enabling the research community to design more generalizable models for this challenging task. Wang~\etal~\cite{wang2023generalizable} proposed a generalizable end-to-end mitosis detection model with the help of an auxiliary nuclei segmentation task and achieved state-of-the-art (SOTA) performance on these two datasets. However, there exists an underlying assumption that there exists a well-trained nuclei segmentation model to help obtain pixel-level annotation of mitosis. This assumption restricts the clinical practicability of the model because pursuing large among of nuclei labels is also time-consuming and challenging. By summarizing the previous studies, we can easily find that existing studies solve the aforementioned challenges mostly in the technical aspect by introducing various novel techniques or increasing the model capacity, such as multi-stage design~\cite{sohail2021mitotic}, attention mechanism~\cite{li2022domain}, multi-scale feature representation~\cite{wang2023generalizable}, ensemble model~\cite{tellez2018whole} and etc. However, they lack consideration of two prerequisites which we believe play the key roles in mitosis detection, a balanced dataset and a diverse feature representation.

In this paper, we rethink mitosis detection in data and feature representation aspects. First of all, since mitosis detection is a needle-in-the-haystack task. The data is extremely imbalanced, not only the quantity, but also the diversity. So a diverse and balanced dataset is the first prerequisite for model training. Second, since the mitotic and non-mitotic cells are heterogeneous. Defining only 2 classes (mitotic or non-mitotic) may not be able to form an informative and diverse feature representation to represent all the morphological appearances. Therefore, a diverse feature representation is the second prerequisite for better performance. To this end, we propose a multi-stage and classification-based approach (MitDet) with point annotations for generalizable mitosis detection. Inspired by the knowledge that mitosis is a biological behavior of nuclei, a hematoxylin-based approach~\cite{zhao2020triple} is first introduced for annotation-free nuclei localization. And then we propose diversity-guided sample balancing (DGSB) to balance the quantity and diversity of the training samples. To tolerance the color variance across different institutions, we propose stain enhancement (SE) to augment the domain-relevant data by our previously proposed RestainNet~\cite{zhao2022restainnet} and enhance the cross-domain feature representation by EFDMix~\cite{zhang2022exact}. Finally, an inter- and intra- class diversity-preserved mitosis classification (InCDP) is proposed to solve the heterogeneity of the morphological appearances by dividing mitosis and non-mitosis into several sub-classes, leading to a more diverse feature representation and more precise detection results.

Extensive experiments have been conducted to evaluate the effectiveness and generalizability of our proposed MitDet with and without domain discrepancy on five popular datasets, including MIDOG2021, AMIDA2013, MITOSIS2014, TUPAC2016, MIDOG2022. MitDet with point annotations trained on MIDOG2021 outperforms not only point-based models but also bbox-based (bounding box) and pixel-level annotations models on both internal validation set (MIDOG2021) and other four external datasets. It demonstrates superior model generalizability while minimizing the annotation efforts for pathologists. Ablation studies show that the rethinking of data balancing and feature representation diversifying are meaningful and effective for mitosis detection in qualitative and quantitative experiments. The main contributions of this study are summarized as follows.

\begin{itemize}
  \item We propose a novel and generalizable mitosis detection framework (MitDet) that achieves SOTA performance on five popular datasets internally and externally comparing with existing models.
  \item We are the first one to consider balancing and diversifying data as well as feature representation, which brings new insights for this task. DGSB is proposed to balance the data quantity and diversity. InCDP is proposed to obtain a diverse feature representation. Using SE module to enhance the diversity of both data and features simultaneously.
  \item MitDet is the one with minimal annotation efforts by point annotations while achieving the best detection performance among all the existing methods, including bbox and pixel annotation-based models.
\end{itemize} 
\section{Related Works}\label{sec2}
\subsection{Mitosis Detection}
Mitosis detection is a challenging task in computational pathology due to the needle-in-a-haystack nature. Various methods have been proposed to formulate it as an object detection task~\cite{ccayir2022mitnet,wang2022novel}, a semantic segmentation task~\cite{li2019weakly,wang2023generalizable} or an image classification task~\cite{sohail2021mitotic,tellez2018whole}, which derive two mainstream pipelines, single-stage ones and multi-stage ones. It remains an open problem since a lot of problems remain unsolved, such as precision, model generalizability, real-world practicability and etc.

\textbf{Single-stage models} typically regard mitosis detection as semantic segmentation or object detection tasks $f_{seg/det}$ in an end-to-end manner. 
\begin{equation}
Y_{pre}=f_{seg/det}(I)
\label{rethink1}
\end{equation}
Li~\etal~\cite{li2019weakly} introduced a concentric circle loss on the basis of the FCN~\cite{long2015fully} segmentation network, which allows their model to obtain segmentation results with only point annotations. However, they did not address the data imbalance problem. Han~\etal~\cite{han2021contextual} utilized spatial area-constrained loss and relied solely on point annotations. Although this model incorporated targeted loss functions to alleviate sample imbalance. It did not consider how to handle the domain discrepancy leading to unsatisfactory model generalizability. Wang~\etal~\cite{wang2023generalizable} handled domain discrepancy by a Fourier-based~\cite{yang2020fda} data augmentation method. Then they used the focal loss to solve the imbalanced data and employed an attention mechanism to make the model focus more on the complex shapes of mitosis. Thanks to the pixel-level pseudo-labels generated by a nuclei segmentation HoVer-Net~\cite{graham2019hover}, their model achieves SOTA performance on multi-center datasets. However, this model heavily relies on the results generated by HoVer-Net. This underlying assumption restricts the clinical practicability of the model since pursuing large-scale nuclei annotations is also expensive and time-consuming.

\textbf{Multi-stage models} usually include two main stages~\cite{ccayir2022mitnet}. The first one is to search for mitosis candidates by a segmenation or a detection model $f_{seg/det}$. And the second one is to classify the selected samples into mitotic cells or non-mitotic cells by a classification model $f_{cls}$.
\begin{equation}
Y_{pre}=f_{cls}(f_{seg/det}(I))
\label{rethink2}
\end{equation}
Compared with one-stage methods, the multi-stage strategy can easily alleviate the data imbalanced problem by balancing the negative samples while a specific classifier may achieve better performance on differentiating mitosis. However, the multi-stage approaches have to ensure a high enough sensitivity of the first stage. Mohmood~\etal~\cite{mahmood2020artificial} used faster RCNN~\cite{ren2015faster} to quickly obtain candidate cells, and then performed classification by fusing the scores from two CNNs. Sohail~\etal~\cite{sohail2021mitotic} applied Mask-RCNN~\cite{he2017mask} to identify candidate patches and then categorized these patches by an ensemble model with five CNN branches. It can be easily observed that current multi-stage models improve the mitosis detection mainly in the technical aspect by either applying more recent object detectors in the first stage or increasing the capacity of the classification model in the second stage. However, the above-mentioned approaches barely consider the data characteristics which may lead to poor model generalizability.

In this study, we comprehensively investigate the data characteristics of the mitosis detection task and consider it mainly from two new perspectives. (1) From the data perspective, we observe that the mitotic cells and non-mitotic cells are highly heterogeneous. Therefore, we not only balance the data quantity but also the diversity. (2) From the feature representation perspective, we believe the binary classification task is not able to represent highly heterogeneous morphological appearances. Therefore, we divide it into several subspaces in order to achieve more diverse feature representation and stronger generalization ability for the model. 

\subsection{Domain generalization for medical image}
Domain generalization (DG) is crucial for deep learning models to recognize the out-of-distribution (OOD) data~\cite{zhou2022domain}. Due to varying imaging acquisition protocols, imaging device vendors and patient populations, domain discrepancy is a common problem encountered in all the medical image modalities~\cite{li2020domain}. Typical DG methods in medical image include data augmentation~\cite{wang2023generalizable,zhang2020generalizing}, meta learning~\cite{li2022domaingeneralization,HAN2022102481} and domain-invariant feature learning~\cite{li2022domain,zhao2021robust}. 

Before 2021, most of the studies mainly focus on how to improve the detection performance rather than model generalizability due to the small-scale datasets. Tellez~\etal~\cite{tellez2018whole} proposed a mitosis detector with invariant staining, which enhances data by directly changing the H\&E color channels, thereby improving the generalization of the model. In recent years, two challenges MIDOG2021 and MIDOG2022 introduce large-scale datasets scanned by different digital slide scanners from different types of tumors, which allow people to pay more attention to the research on domain generalization in the mitosis detection task. In these two competitions, participants use various methods to overcome domain gaps in mitosis. Data augmentation and domain-invariant feature learning are common techniques. Li~\etal~\cite{li2022domain} proposed a method to learn domain-independent features by aligning different domains through the cross-domain adaptive module (CDAM), which was trained in an adversarial manner. However, this model focused on domain adaptation by fast adapting the model to the target domain with fewer samples. It has no specific design to support domain generalization on the unseen dataset. Wang~\etal~\cite{wang2023generalizable} relied on a fourier-based data augmentation method to improve the model generalizability. However, the fourier-based method is not able to learn domain specific distribution while has to select an appropriate parameter. 

By investigating the characteristics of mitosis, the domain discrepancy mainly comes from color variation introduced by different staining reagents or digital slide scanners. Another factor that might lead to domain gaps is the morphological differences across different tumors. However, since mitosis is a biological behavior in all the eukaryotic cells which shares exactly the same four phases. We believe that the morphological discrepancy is less influential than the color variation on domain generalization. In this study, we alleviate domain discrepancy by first introducing unsupervised stain augmentation using our proposed RestainNet~\cite{zhao2022restainnet} to introduce domain specific samples in an unsuperivsed learning manner and then applying EFDMix~\cite{zhang2022exact} to enhance the cross-domain feature representation. 

\section{Method}\label{sec3}
In this section, we systematically present our proposed MitDet. We first show how we rethink the mitosis detection task and the insights of the model in Section~\ref{3.1}. Then we demonstrate each step of our framework, including hematoxylin-based nuclei localization in Section~\ref{3.3}, diversity-guided sample balancing (DGSB) in Section~\ref{3.4}, Stain Enhancement (SE) in Section~\ref{3.5} and inter- and intra-class diversity-preserved mitosis classification (InCDP) in Section~\ref{3.6}. Section~\ref{3.7} shows the implementation details. 

\subsection{Rethinking of mitosis detection}\label{3.1}
We first comprehensively analyze the challenges of mitosis detection by systematically investigating the characteristics of mitosis and demonstrating the insights of our proposed model. 

\subsubsection{Annotations}
Currently, there are three common labeling ways, including pixel-level, box-level and point-level annotations (Fig.~\ref{fig:rethink}~(a)). However, pursuing large-scale annotations is extremely time-consuming due to the needle-in-a-haystack nature and expertise dependence. Point annotation is the easiest labeling way but finer annotations will bring stronger supervision for model training. It is a trade-off between the annotation efficiency and the information richness of the labels. To alleviate the trade-off, we leverage the biological and chemical knowledge that mitosis is a biological behavior of the cell division which only happens in the nuclei and hematoxylin stains nuclei to show blue. According to the knowledge, we introduce a hematoxylin-based nuclei localization to pre-locate the candidates of the mitosis with high sensitivity. For each candidate, point annotations are enough to help us define whether a nucleus is a mitosis or not.

\subsubsection{Imbalance}
Since mitosis is an extremely rare event (Fig.~\ref{fig:rethink}~(b)), accounting for around 0.3\% proportion of all the cells in MIDOG2021. Conventional approaches deal with the imbalanced problem by resampling the dataset or introducing a proper loss function. However, existing approaches only consider the imbalanced problem of data quantity. Due to the heterogeneity of mitosis, data diversity and difficulty should also be balanced. Otherwise, imbalanced hard negative samples may lead to poor detection performance. In order to alleviate the imbalanced problem, we propose a diversity-guided sample balancing approach to balance the data quantity, diversity and difficulty.

\subsubsection{Color Discrepancy}
Color variation, shown in Fig.~\ref{fig:rethink}~(c), is one of the major reasons that affect the model's generalizability in the mitosis detection task. Different staining reagents, protocols or digital slide scanners may introduce color variations. Stain normalization is the most common way for maintaining color consistency. However, conventional stain normalization approaches have to choose an appropriate reference image and may generate some artifacts while damaging the image information. In this study, we solve color discrepancy in both data and feature representation levels by augmenting the color domains and enhancing the cross-domain feature representation simultaneously. 

\subsubsection{Heterogeneity}
\begin{figure}
	\includegraphics[width=.99\linewidth]{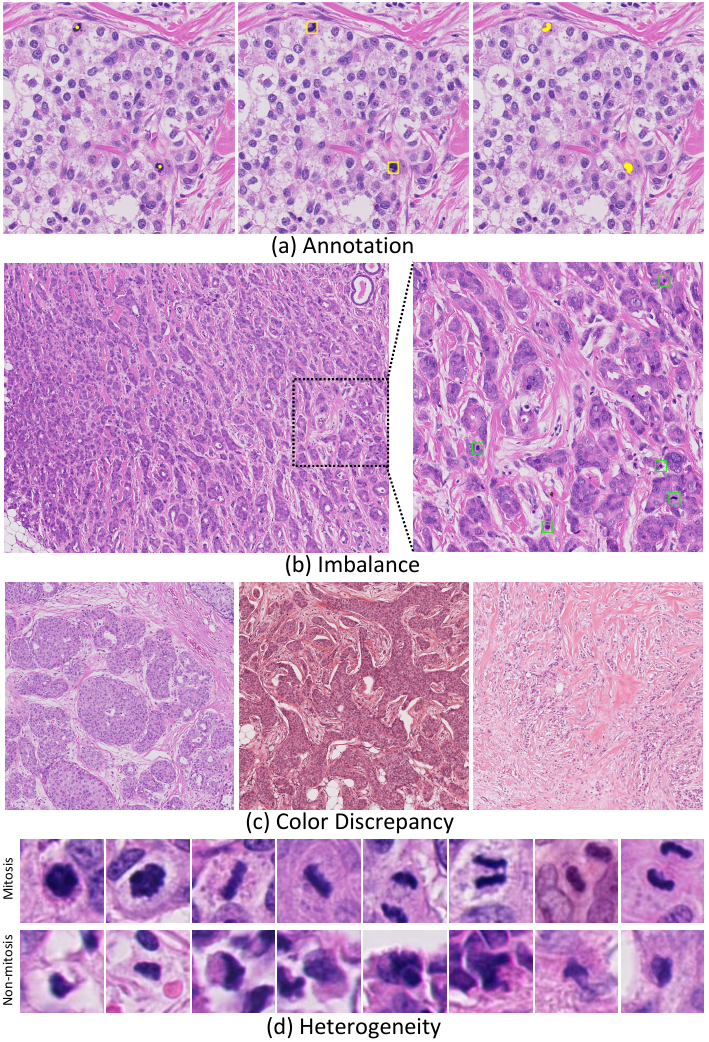}
	\caption{Four major challenges of mitosis detection.}
	\label{fig:rethink}
\end{figure}

What makes mitosis detection so challenging is the heterogeneity (Fig.~\ref{fig:rethink}~(d)). Since mitosis is a dynamic process of cell division. Besides the four phases of mitosis (prophase, metaphase, anaphase, and telophase), there are also some intermediate appearances between two typical phases which are quite ambiguous. Furthermore, apoptotic nuclei, polymorphic tumor epithelial nuclei or even fragile cells may also demonstrate similar morphological appearances to mitotic cells. Conventional binary detection/segmentation/classification tasks might not be able to form an informative and diverse feature representation to represent the heterogeneous data. To make feature representation more informative, we propose to further divide the positive and negative classes into more sub-classes by an unsupervised clustering in the feature space. We hope such a strategy can better differentiate the hard samples according to the morphological differences. 

\begin{figure*}
	\includegraphics[width=.99\linewidth]{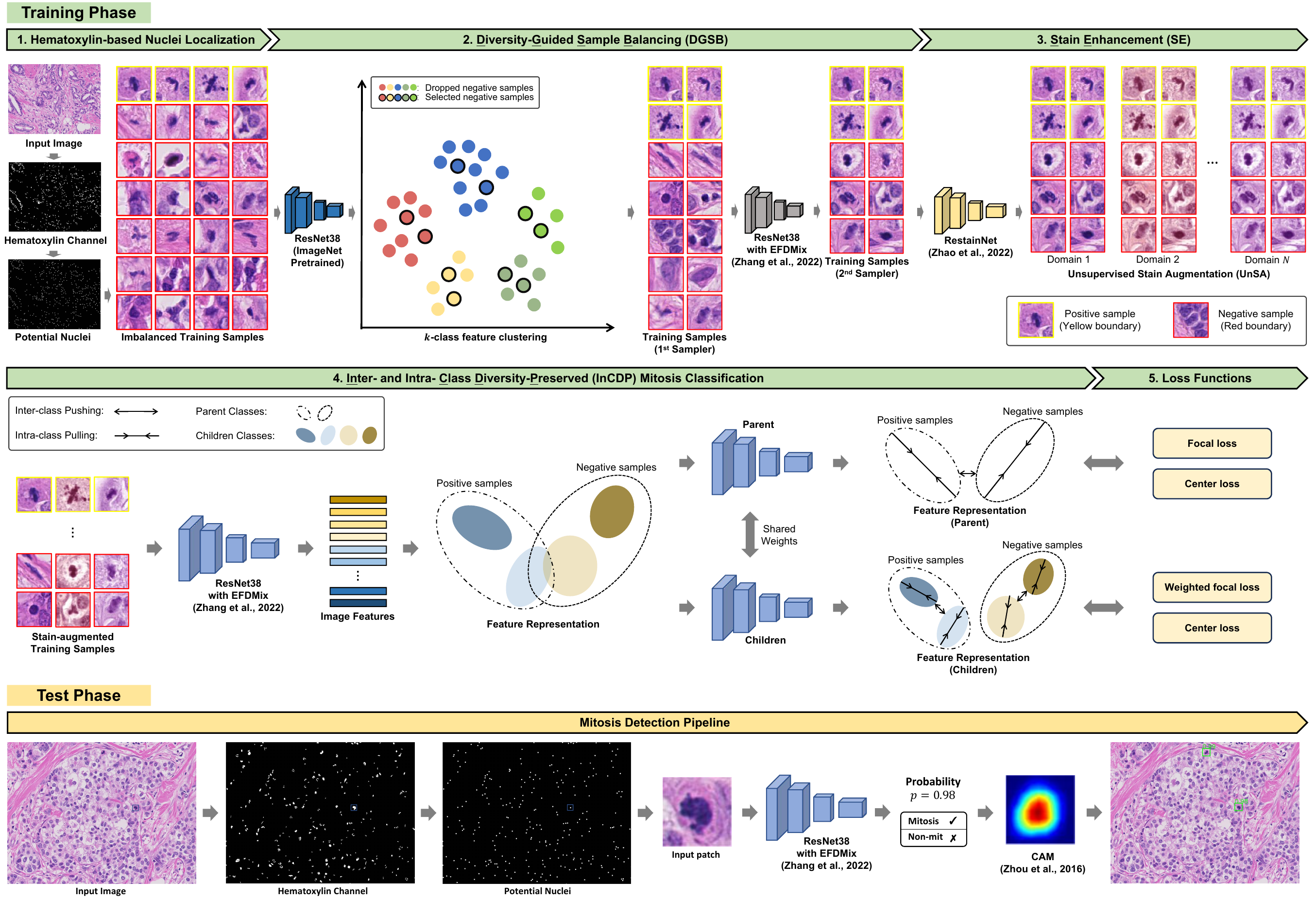}
	\caption{Framework overview of the proposed MitDet. Hematoxylin-based nuclei localization serves for locating all the nuclei by the hematoxylin channel. Diversity-guided sample balancing (DGSB) removes the redundant negative samples and balances the quantity, diversity and difficulty of the training data. Stain Enhancement (SE) with unsupervised stain augmentation (UnSA) and EFDMix is proposed to improve the cross-domain feature representation. Inter- and intra- class diversity-preserved (InCDP) approach leads to a more diverse feature representation by dividing mitosis and non-mitosis into several sub-classes. During the test phase, we classify the patches extracted by hematoxylin-based nuclei localization by the classification model trained in InCDP. CAM is finally applied to provide the location of the mitosis and achieve the detection results.}
	\label{fig:workflow}
\end{figure*}

\subsection{Hematoxylin-based Nuclei Localization}\label{3.3}
First of all, we introduce a hematoxylin-based nuclei localization approach to roughly identify all the nuclei. Different from existing approaches that rely on mitosis or nuclei annotations. Our method is totally label-free by leveraging biological and chemical knowledge. Mitosis is a biological behavior of the cell division which only happens in the nuclei and hematoxylin stains nuclei to show blue.

\begin{figure}
	\includegraphics[width=.99\linewidth]{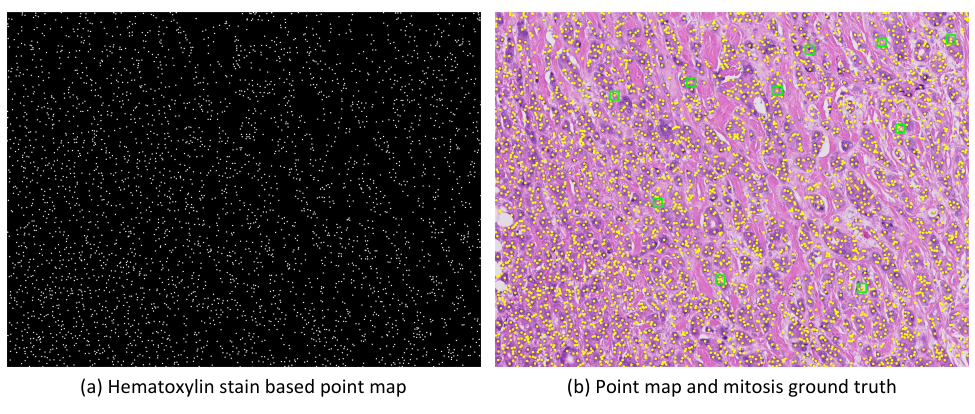}
	\caption{We achieve nuclei localization by the extracted hematoxylin channel. (a) shows the point map transformed from the hematoxylin channel image. (b) shows the locations of nuclei (yellow points) and the ground truth of mitosis (green boxes).}
	\label{fig:h_channel}
\end{figure}

Inspired by our previous study~\cite{zhao2020triple}, hematoxylin stain can guide the nuclear segmentation task to achieve clearer nuclei boundaries and could be used for nuclear localization. Specifically, we apply a color decomposition technique~\cite{ruifrok2001quantification} to extract the hematoxylin channel from the RGB image $I$. By extracting the centroid coordinates of each nucleus, we can crop patches $D=\{I_1,I_2,...,I_l\}$ for the following image classification. We divide the sampled images into positive samples $D^P$ and negative samples $D^N$, where the positive samples contain mitosis that is present in the image. There are two advantages of hematoxylin-based nuclei localization. First, it does not require any additional annotation or training process. Second, due to the optical principle of color decomposition, the hematoxylin channel is insensitive to different color domains and can provide accurate nuclei position. Fig.~\ref{fig:h_channel} demonstrates an example of the hematoxylin channel and the extract point map of the nuclei and mitosis.

\subsection{Diversity-Guided Sample Balancing}\label{3.4}
Different from existing studies that balance the negative (non-mitosis) samples with random sampling~\cite{sohail2021mitotic,mahmood2020artificial,ccayir2022mitnet}, we achieve it in three perspectives, including quantity, diversity and difficulty. The proposed Diversity-Guided Sample Balancing (DGSB) approach contains two samplers, $1^{st}$ sampler for balancing quantity and diversity, $2^{nd}$ sampler for balancing difficulty. DGSB maintains a balanced, diverse and informative training set while enabling the model to learn features more efficiently and stably.

\begin{figure}
	\includegraphics[width=.99\linewidth]{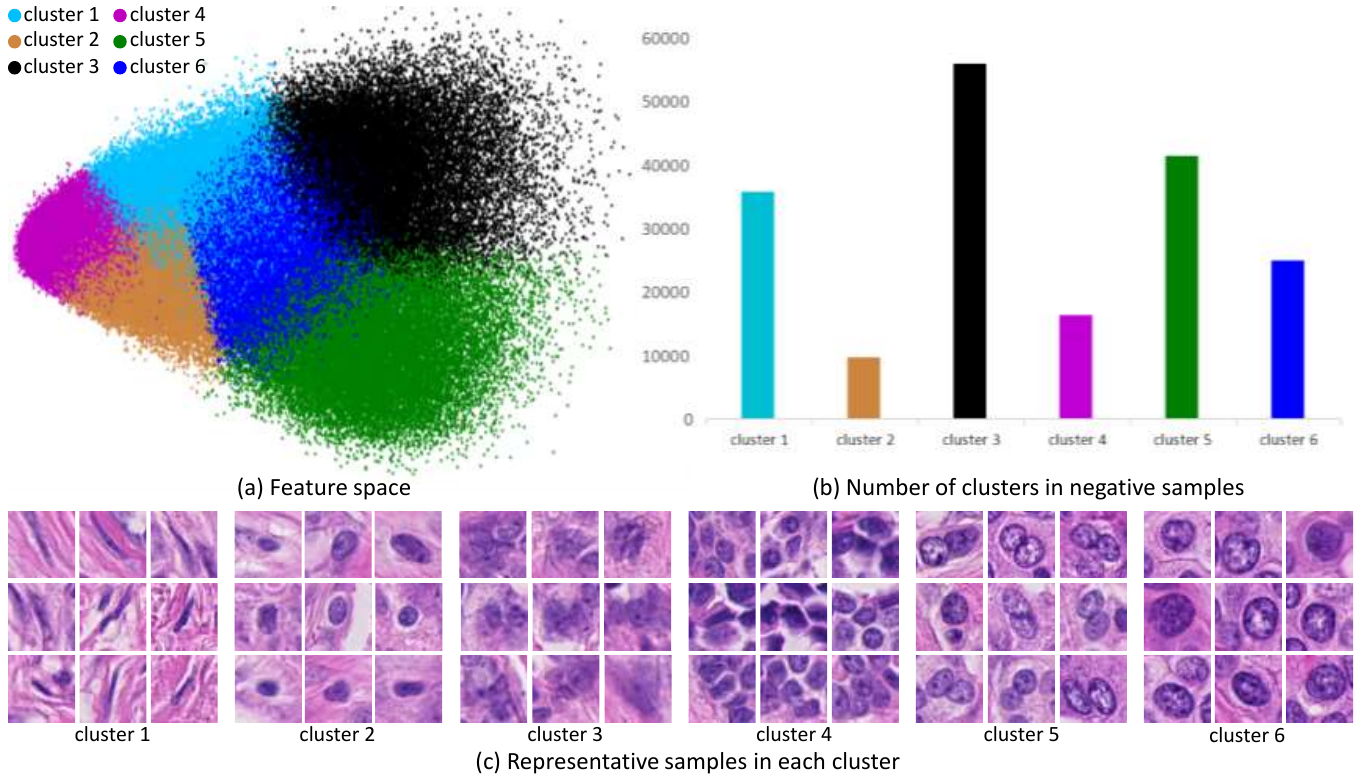}
	\caption{Visualization of negative sample clustering. (a) Feature space of the clustering. (b) The number of patches in each cluster. (c) Some selected samples in each cluster.}
	\label{fig:1-sampler}
\end{figure}

\textbf{$1^{st}$ Sampler (Quantity and diversity)}:
Given the negative patch samples $D^N$ extracted by the previous hematoxylin-based nuclei localization, we first extract the feature vectors of the patches by an ImageNet pre-trained classification model, ResNet38~\cite{wu2019wider}.  
Then we use k-means clustering to divide the negative samples into $k$ clusters such that $D^N=\{C_1,C_2,...,C_k\}$. Fig.~\ref{fig:1-sampler} demonstrates (a) the feature space, (b) the total number and (c) the selected samples of each cluster. We can easily observe great differences in the morphological appearances and the numbers among different clusters. DGSB randomly samples the same among negative patches from each cluster to alleviate the imbalanced problem in both quantity and diversity, shown as follows:
\begin{equation}
	D_{1^{st}}^{N}=\{C_1^{'},C_2^{'},...,C_k^{'}\}
\end{equation}

\textbf{$2^{nd}$ Sampler (Difficulty)}: 
By our observation, even though the quantity and diversity of the data are balanced, the hard negative samples are still not well differentiated since they only account for a very small amount. We can also observe that most of the samples shown in Fig.~\ref{fig:1-sampler}~(c) are obviously different from mitosis in morphology. Therefore, we introduce a $2^{nd}$ sampler to further balance the difficulty of the negative samples.

\begin{equation}
	f_{\rm diff}(\{D_{1^{st}}^{N},D^P\},\theta)
\end{equation}

We train a classification model $f_{\rm diff}$ using the sampled negative patches $D_{1^{st}}^{N}$ and all the positive patches $D^P$. For each negative sample $i$, it will be dropped if the model shows high confidence with the predicted probability less than $\epsilon=0.5$. This strategy can filter out most of the easy samples and the hard ones are left. After the $2^{nd}$ sampler, we can now obtain balanced dataset $\mathbf{D}=\{D_{2^{nd}}^N, D^P\}$.

\subsection{Stain Enhancement}\label{3.5}
Since the domain discrepancy of the mitosis detection task mainly comes from stain or color variance. Therefore, on the basis of a balanced dataset, we further enhance the stain by an unsupervised stain augmentation (UnSA) in the data aspect and an EFDMix in the feature representation aspect.

\textbf{Unsupervised Stain Augmentation}:
In the data aspect, we want to enrich the information of the dataset rather than reduce or manipulate it. Thus, different from conventional approaches that apply stain normalization to normalize all the images into one color domain. We augment the training data into multiple color domains by applying our previous study, RestainNet~\cite{zhao2022restainnet}. Unlike conventional stain normalization approaches that may generate image artifacts when an inappropriate reference image is selected. RestainNet can well preserve the structural information by learning a specific color domain using large-scale unlabeled data in an unsupervised manner. As shown in Fig.~\ref{fig:workflow}, we augment the training data $\mathbf{D}$ into $n$ color domains and form the final training data $\mathbf{\hat {D}}$.
\begin{equation}
	\mathbf{\hat {D}}=\{\mathbf{D},\mathbf{D}_{1},\mathbf{D}_{2},...,\mathbf{D}_{n}\}
\end{equation}

In practice, we augment three common domains from different medical centers or digital slide scanners. We can further expand as many domains as we want with no annotation burden if it is necessary.

\textbf{EFDMix}:
In the feature representation aspect, we also want the model to learn a more robust and comprehensive cross-domain feature representation. Inspired by a style transfer method that augments cross-style features by mixing feature distributions, we apply an EFDMix~\cite{zhang2022exact} method to augment the feature representation of different stain styles. Given two extracted features $U=(u_1,u_2,...,u_n)$ and $V=(v_1,v_2,...,v_n)$ from two images from different color domains, we fuse the domain knowledge by mixing two feature distributions as follows:
\begin{equation}
	{\rm EFDMix}(U,V):w_{i}=u_{i}+(1-\mu)v_{i}-(1-\mu)\langle u_{i}\rangle
	\label{EFDMix_2}
\end{equation}
where $\langle .\rangle$ represents the stop-gradient~\cite{chen2021exploring} operation. Instance-wise mixing weight $\mu$ is sampled from a Beta-distribution. By introducing stain-augmented data and mixing cross-domain feature distributions, the model generalizability can be improved.

\subsection{Inter- and Intra- Class Diversity-Preserved Mitosis Classification}\label{3.6}
Even though we have already balanced the dataset, distinguishing mitotic cells from non-mitotic cells is still a challenging task since both of them demonstrate complex morphological features. There exists a large intra-class variety and a high inter-class similarity which hinder the conventional binary classification task to form a diverse feature representation. To tackle this challenge, we propose an inter- and intra- class diversity-preserved (InCDP) mitosis classification model.

\subsubsection{Generation Of Child Classes}
Besides the binary classification task to distinguish mitotic and non-mitotic cells, we further expand each class into $T$ child classes to encourage the model to focus on not only inter-class differences but also intra-class differences, finally obtaining a more diverse feature representation. 

\begin{align*}
	\begin{split}
		Y_{p} \rightarrow Y_{c}= \left \{
		\begin{array}{ll}
			(0,...,0,Y_{c}^{T+1},..., Y_{c}^{2T}), & Y_{p} = 1\\
			(Y_{c}^1,..., Y_{c}^{T},0,...,0), & Y_{p} = 0\\
		\end{array}
		\right.
	\end{split}
\end{align*}
where $Y_{p}\in \{0,1\}$ is the parent label, $Y_{c}\in\{0,1\}^{2T}$ is the one-hot child label. 

We first train a binary classification model ResNet38 with EFDMix using the parent class labels until the model converges. Then we extract the deep features of each sample from the parent model. After that, the pseudo-labels of the child classes are obtained by unsupervised k-means clustering.

\subsubsection{Joint Training with Child and Parent Classes}
Now we jointly train the model by leveraging both parent and child classes. The objective of the classification model is to distinguish not only parent classes but also child classes. Since the mitosis detection task has a large intra-class variety and a high inter-class similarity. Therefore, for both parent and child classes, we introduce inter-class pushing to better differentiate different classes, and intra-class pulling to shrink the feature representation for the samples with the same class. To achieve this, we introduce focal loss and center loss for both parent and child classes at the same time. 

\textbf{For parent classes:}
Focal loss is introduced to push away different classes in the feature space by down-weighting the contribution of the easy examples and focusing more on the hard ones. The focal loss function is given by:
\begin{equation}
	L_{Focal}^{p}=-\sum\limits_{i}[(1-\hat{y_i})^\gamma y_i log\hat{y_i}+(\hat{y_i})^\gamma (1-y_i)log(1-\hat{y_i})]
	\label{focal_1}
\end{equation}
where $y_{i}$ and $\hat{y_{i}}$ denote the ground truth and predicted label of the $i^{th}$ sample respectively, and $\gamma$ is a tunable parameter that controls the weight assigned to the misclassified examples. 

Center loss is to minimize the intra-class variability by encouraging the feature vectors of the same class to be close to their corresponding class center. The center loss function is defined as:
\begin{equation}
	\mathcal{L}_{Center}^p=\frac{1}{2} \sum\limits_{i=1}\limits^{N} {\Vert x_i-c_{y_i} \Vert_2^{2}}
	\label{center_1}
\end{equation}
where $x_i$ is the feature vector of the $i^{th}$ sample, $y_i$ is its corresponding class label, $N$ is the number of samples, and $c_{y_i}$ is the center of the ${y_i}^{th}$ class.

\textbf{For child classes:}
We introduce a weighted focal loss to specifically handle the hard samples by introducing a center distance weight $\omega$. The samples from the child class with a longer distance to the opposite class will be regarded as easier samples, resulting in a lower weight.
\begin{equation}
	\mathcal{L}_{Focal}^{c}=-\sum\limits_{j} w_j \sum\limits_{i}[(1-\hat{y_{ji}})^\gamma y_{ji} log\hat{y_{ji}}]
	\label{focal_2}
\end{equation}
where $y_{ji}$ and $\hat{y_{ji}}$ denote the ground truth and predicted label of the $i^{th}$ sample in the $j^{th}$ child class. In addition, child classes use the same center loss as the parent classes shown in Eq.~\ref{center_1}.

The parent and child losses are used jointly to optimize the classification model.
\begin{equation}\label{joint_loss}
	\mathcal{L}_{total}=\mathcal{L}_{Focal}^p + \mathcal{L}_{Center}^p +\lambda (\mathcal{L}_{Focal}^c + \mathcal{L}_{Center}^c)
\end{equation}
where $\lambda$ is the weight used to balance the losses of the parent and child classes.

\subsection{Implementation Details}
\label{3.7}
\textbf{Test phase:} We first use the hematoxylin-based nuclei localization method to roughly identify the center points of all the nuclei. For each center point, we crop a fixed-size patch. The classifier trained by InCDP is used to predict whether the patch contains mitotic cells. Finally, class activation maps are applied to provide the location of mitosis.

Our model is optimized using the SGD optimizer with a weight decay of 5e-4. The learning rate is 1e-3. The mini-batch size for model training is set to 150. Our experiments are carried out on the PyTorch framework and a workstation equipped with a 24 GB memory NVIDIA RTX3090 GPU. We set the size of the patch to be $80*80$. In the DGSB module, we use pre-trained ResNet on ImageNet and K-means to divide the negative samples into 10 clusters. In UnSA, We extend the stain color to 3 color spaces. In InCDP, we use K-means to divide both the negative and positive samples into 4 clusters respectively, and set $\lambda=0.5$ for the loss function. Real-time data augmentation is applied to increase the diversity of the training data, which includes random horizontal, vertical, and 90-degree flipping, random elastic, random brightness, contrast, HED color space-changing and Random Gaussian blur.

\section{Experimental Results}\label{sec5}
Our proposed MitDet is evaluated on the following five datasets, including MIDOG2021\footnote{https://midog2021.grand-challenge.org/}, AMIDA2013~\cite{veta2015assessment}, MITOSIS2014\footnote{https://mitos-atypia-14.grand-challenge.org/Dataset/}, TUPAC2016\footnote{https://tupac.grand-challenge.org/Dataset/}, and MIDOG2022\footnote{https://midog2022.grand-challenge.org/} with the same experimental configurations with~\cite{wang2023generalizable}. Our method was trained on the MIDOG2021 dataset and evaluated on the internal dataset of MIDOG2021 and the other four external datasets. We first compare our method with the state-of-the-art(SOTA) on the internal dataset. And then we evaluate the model generalizability on the external datasets. Finally, we validate the effectiveness of the novelties we proposed in the ablation experiments.

\begin{table}[htp]
	\small
	\centering
	\caption{Comparison with the existing approaches on the internal dataset of MIDOG2021.}
	\label{table_sota}
	\begin{threeparttable}
		\begin{tabular}{cc|ccc}
			\toprule
			\textbf{Methods} &\textbf{Label} &\textbf{F1} &\textbf{Recall} &\textbf{Prec} \\ 
			\midrule
			DeepMitosis\tnote{2}~\cite{li2018deepmitosis} &pixel &0.6440 &0.6265 &0.6624\\
			MaskMitosis\tnote{1}~\cite{sebai2020maskmitosis} &pixel &0.6601 &0.6024 &0.7299\\
			DHE-Mit\tnote{2}~\cite{sohail2021mitotic} &pixel &0.6996 &0.7200 &0.6804\\
			FMDet\tnote{1}~\cite{wang2023generalizable} &pixel &0.7773 &\textbf{0.8146} &0.7440\\ \midrule
			SegMitos\tnote{1}~\cite{li2019weakly} &point &0.6791 &0.6271 &0.7405\\
			RetinaNet\tnote{1}~\cite{aubreville2021quantifying} &box &0.6236 &0.6687 &0.5842\\
			Ours\tnote{2} &point &\textbf{0.7895} &0.7715 &\textbf{0.8084}\\
			\toprule
		\end{tabular}
		\begin{tablenotes}    
			\footnotesize               
			\item[1] Single-stage methods.          
			\item[2] Multi-stage methods.        
		\end{tablenotes} 
	\end{threeparttable}
\end{table}

In the experiments, We compare MitDet with six SOTA mitosis detection methods using different annotation ways, including \textbf{pixel-level} methods DeepMitosis~\cite{li2018deepmitosis}, MaskMitosis~\cite{sebai2020maskmitosis}, DHE-Mit~\cite{sohail2021mitotic}, and FMDet~\cite{wang2023generalizable}, the \textbf{box-level} method RetinaNet~\cite{aubreville2021quantifying}, and the \textbf{point-level} method SegMitos~\cite{li2019weakly}, as well as some baseline models~\cite{wollmann2017deep,tellez2018whole,lei2020attention,li2022domain}.

\subsection{Comparison with state-of-the-art methods on internal dataset}
We first compare the model capability on the internal dataset of MIDOG2021 in Table~\ref{table_sota}. The same with existing studies, we train the model on the training set of MIDOG2021. As we can see, our proposed method generally outperforms all other competitors in terms of F1 score as well as precision and achieves the second-highest recall. It not only demonstrates over 10\% higher detection performance than the two weak annotation approaches but also pixel-level ones. Even compared to the previous SOTA method (FMDet), MitDet still outperforms it by around 1\% in terms of F1 score and over 5\% in terms of precision. Besides the detection performance, MitDet demonstrates great efficiency by requiring point annotations only. However, the SOTA approach FMDet relies on a nuclei segmentation model trained on a large-scale pixel-level nuclei dataset to generate pixel-level mitosis annotations. MitDet greatly saves the annotation burden for pathologists. Furthermore, our model demonstrates great capability and flexibility only using ResNet38 without any complex model design or postprocessing step.
Notice that, due to the robustness of the model, it is not necessary for our model to specifically select an appropriate threshold for each dataset.

We believe that rethinking mitosis detection on the data and the feature perspectives are a milestone for this task. Realizing the staining characteristics helps us reduce the annotation efforts with point labels only. For the first time, balancing the diversity, quantity and difficulty of the training samples helps the model overcome the needle-in-a-haystack challenge. Enhancing the feature representation helps better distinguish the hard negative samples, while greatly reducing the false positive rate.

\begin{table}[htp]
	\centering
	\scriptsize
	\caption{Comparison with existing approaches on external datasets.}
	\label{table_domain}
	\begin{threeparttable}
		\begin{tabular}{ccc|ccc}
			\toprule
			\textbf{Datasets} &\textbf{Methods} &\textbf{Label} &\textbf{F1} &\textbf{Recall} &\textbf{Prec} \\
			\midrule
			\multirow{4}{*}{\makecell{AMIDA\\2013}} &Wollmann~\etal\tnote{1}~\cite{wollmann2017deep} &point &0.6090 &0.6860 &0.5470 \\ 
			&SegMitos\tnote{1}~\cite{li2019weakly}        &point &0.6728 &0.6773 &\textbf{0.6685} \\
			&FMDet\tnote{1}~\cite{wang2023generalizable}           &pixel &0.6791 &\textbf{0.7405} &0.6271 \\
			&Ours\tnote{2} &point &\textbf{0.6832} &0.7254 &0.6457 \\
			\midrule
			\multirow{5}{*}{\makecell{MITOSIS\\2014}} &DeepMitosis\tnote{2}~\cite{li2018deepmitosis} &pixel &0.4370 &0.4430 &0.4310 \\
			&Lei~\etal\tnote{2} ~\cite{lei2020attention} &pixel &0.4000 &/ &/ \\
			&MaskMitosis\tnote{1}~\cite{sebai2020maskmitosis}        &pixel &0.4750 &0.4530 &\textbf{0.5000} \\
			&FMDet\tnote{1}~\cite{wang2023generalizable}              &pixel &0.4901 &0.5564 &0.4380 \\
			&Ours\tnote{2}     &point &\textbf{0.5139} &\textbf{0.6011} &0.4488 \\
			\midrule
			\multirow{7}{*}{\makecell{TUPAC\\2016}} &Lafarge~\etal\tnote{1}~\cite{lafarge2017domain} &point &0.6200 &/ &/ \\
			&Akram~\etal\tnote{2}~\cite{akram2018leveraging} &point &0.6400 &0.6710 &0.6130 \\
			&Tellez~\etal\tnote{2}~\cite{tellez2018whole} &point &0.4800 &/ &/ \\
			&SegMitos\tnote{1}~\cite{li2019weakly}             &point &0.6740 &0.7544 &0.6091 \\
			&DBIN\tnote{1}~\cite{li2022domain}                 &box &0.7390 &0.8130 &0.6780 \\
			&FMDet\tnote{1}~\cite{wang2023generalizable}       &pixel &\textbf{0.7458} &\textbf{0.8018} &0.6971 \\
			&Ours\tnote{2}                           &point &\textbf{0.7458} &0.7333 &\textbf{0.7586} \\
			\midrule
			\multirow{6}{*}{\makecell{MIDOG\\2022}} &DeepMitosis\tnote{2}~\cite{li2018deepmitosis} &pixel &0.6375 &0.7656 &0.5462 \\
			&SegMitos\tnote{1}~\cite{li2019weakly} &point &0.6797 &0.7603 &0.6146 \\
			&RetinaNet\tnote{1}~\cite{aubreville2021quantifying}  &box &0.6145 &0.7325 &0.5292 \\
			&DHE-Mit\tnote{2}~\cite{sohail2021mitotic}  &pixel &0.6863 &0.7524 &0.6309 \\
			&FMDet\tnote{1}~\cite{wang2023generalizable}   &pixel &0.7389 &0.7796 &0.7022 \\
			&Ours\tnote{2} &point &\textbf{0.7519} &\textbf{0.7840} &\textbf{0.7223}\\
			\toprule
		\end{tabular}
		\begin{tablenotes}    
			\footnotesize               
			\item[1] Single-stage methods.          
			\item[2] Multi-stage methods.        
		\end{tablenotes} 
	\end{threeparttable}
\end{table}

\subsection{Comparison with related methods on external datasets}
We evaluate the generalizability of MitDet by directly inferring the trained model using MIDOG2021 training set on four external datasets, including AMIDA2013, MITOSIS2014, TUPAC2016 and MIDOG2022. Most of the quantitative results of the existing methods are obtained from the corresponding publications. We keep the same experimental configurations for a fair comparison. Single-stage and multi-stage approaches are distinguished by the superscript $^1$ and $^2$.

Table~\ref{table_domain} demonstrates the quantitative results which were calculated on completely independent and unseen test sets. As can be seen, our model achieves SOTA performance in F1 score on all the external datasets. To be specific, MIDOG2022 is the one with the smallest domain discrepancy but the largest mitosis quantity. MitDet achieves superior performance in all the evaluation metrics F1 score, recall and precision among all the methods. In TUPAC2016, MitDet shows the same F1 score as FMDet, but it demonstrates more balanced recall and precision with a lower false positive rate. Since AMIDA2013 and MITOSIS2014 are two datasets with larger domain discrepancies, the detection performance of existing approaches is obviously lower than the performance of the other two datasets. MitDet improves F1 score by around 2\% on MITOSIS2014 and 1\% on AMIDA2013 which demonstrates the superiority of the model generalizability. 

By analyzing how existing approaches overcome domain discrepancy, we find that most of them rely on normalization approaches to standardize all the images into one specific color domain. The strong baseline model FMDet uses Fourier data augmentation to counter domain shift. However, existing approaches are highly dependent on the selection of the reference image or the augmentation parameter. During the preprocessing step, the image information may lose and the structural information may be harmed due to the image artifacts. Our model can learn more robust domain-specific features using large-scale unlabeled data and augment the staining appearances without losing any structural information. In addition, the unsupervised manner demonstrates great flexibility by allowing the model to extend as many staining domains as we want.

\begin{table}[t]
	\small
	\centering
	\scriptsize
	\caption{Ablation studies.}
	\label{ablationstudy}
	\begin{threeparttable}
		\begin{tabular}{cccccccc}
			\toprule
			\textbf{Datasets} &\textbf{Models} &\textbf{DGSB} &\textbf{SE} &\textbf{InCDP} &\textbf{F1} &\textbf{Recall} &\textbf{Prec} \cr
			\hline
			\multirow{7}{*}{\makecell[c]{MIDOG\\2021\\(val)}}&(1) &- &- &- &0.6993 &0.7473 &0.6572 \cr
			&(2) &$\checkmark$ &- &- &0.7324 &0.7580 &0.7085 \cr
			&(3) &- &$\checkmark$ &- &0.7046 &0.7822 &0.6409 \cr
			&(4) &- &- &$\checkmark$ &0.7179 &0.7526 &0.6862 \cr
			&(5) &$\checkmark$ &$\checkmark$ &- &0.7418 &0.7607 &0.7237 \cr
			&(6) &- &$\checkmark$ &$\checkmark$ &0.7357 &0.7634 &0.7100 \cr
			&(7) &$\checkmark$ &- &$\checkmark$ &0.7880 &\textbf{0.7795} &0.7967 \cr
			&(8) &$\checkmark$ &$\checkmark$ &$\checkmark$ &\textbf{0.7895} &0.7715 &\textbf{0.8084} \cr  
			\hline
			\multirow{7}{*}{\makecell[c]{MIDOG\\2022}}&(1) &- &- &- &0.6697 &0.7069 &0.6363 \cr
			&(2) &$\checkmark$ &- &- &0.6867 &0.7047 &0.6696 \cr
			&(3) &- &$\checkmark$ &- &0.7018 &0.7186 &0.6859 \cr
			&(4) &- &- &$\checkmark$ &0.6873 &0.7193 &0.6587 \cr
			&(5) &$\checkmark$ &$\checkmark$ &- &0.7154 &0.7336 &0.6982 \cr
			&(6) &- &$\checkmark$ &$\checkmark$ &0.7248 &0.7442 &0.7064 \cr
			&(7) &$\checkmark$ &- &$\checkmark$ &0.7368 &0.7706 &0.7059 \cr
			&(8) &$\checkmark$ &$\checkmark$ &$\checkmark$ &\textbf{0.7519} &\textbf{0.7840} &\textbf{0.7223} \cr 
			\hline
			\multirow{7}{*}{\makecell[c]{MITOSIS\\2014}}&(1) &- &- &- &0.2372 &0.2916 &0.2000 \cr
			&(2) &$\checkmark$ &- &- &0.2512 &0.2916 &0.2207 \cr
			&(3) &- &$\checkmark$ &- &0.3179 &0.3690 &0.2792 \cr
			&(4) &- &- &$\checkmark$ &0.2370 &0.2857 &0.2025 \cr
			&(5) &$\checkmark$ &$\checkmark$ &- &0.4161 &0.4579 &0.3814 \cr
			&(6) &- &$\checkmark$ &$\checkmark$ &0.3903 &0.4242 &0.3615 \cr
			&(7) &$\checkmark$ &- &$\checkmark$ &0.3529 &0.3214 &0.3913 \cr
			&(8) &$\checkmark$ &$\checkmark$ &$\checkmark$ &\textbf{0.5139} &\textbf{0.6011} &\textbf{0.4488}\\
			\bottomrule
			\hspace{1mm}
		\end{tabular}
	\end{threeparttable}
\end{table}

\subsection{Ablation study}
We conduct ablation experiments to verify the effectiveness of the proposed novelties, including DGSB, SE and InCDP on the internal validation set of MIDOG2021 and the external datasets MIDOG2022 and MITOSIS2014. We compare the final model with several baseline models with different combinations. Quantitative results are shown in Table~\ref{ablationstudy}.

\begin{figure}[t]
	\centering
	\includegraphics[width=.95\linewidth]{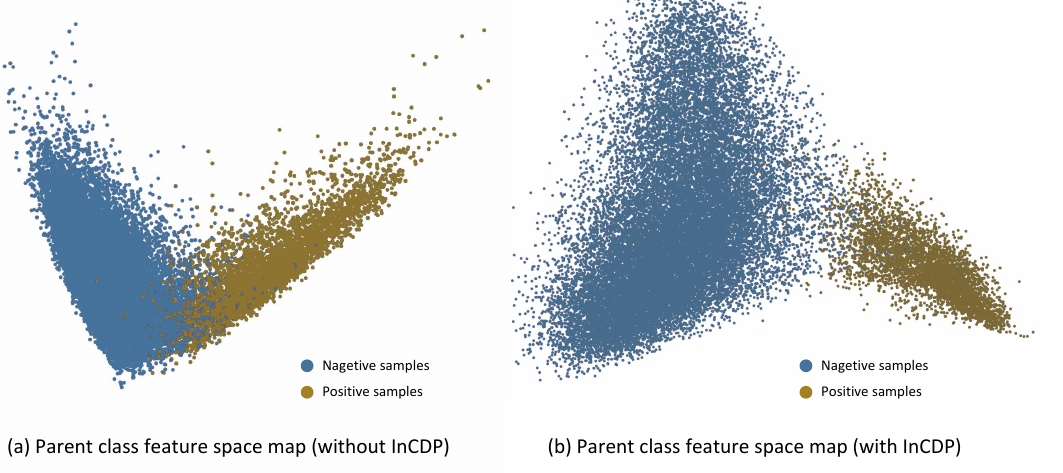}
	\caption{Visualization of the feature spaces with and without InCDP.}
	\label{fig:InCDP}
\end{figure}

On the internal validation set of MIDOG2021, we find that using DGSB (Model (2)) to balance the training data significantly improves the model performance, especially precision. Because DGSB drops a large number of redundant and easy negative samples and maintains the sample diversity, enabling the model to learn more distinguishable features of mitosis and reduce the false positive rate. When applying stain enhancement alone (Model (3)), the performance of the model only shows a slight improvement due to the small domain gap between the validation set and the training set. When equipping InCDP alone (Model (4)), the feature representation of the model is enhanced, leading to a significant improvement. By comparing models (5), (6), (7) with models (2), (3), (4), we can find that associating any two modules together achieves better results. Specifically, Combining DGSB and InCDP improves F1 score from 0.6993 to 0.7895. Although the effectiveness of SE module is limited on the internal validation set, introducing multiple domain data and cross-domain feature enhancement still slightly benefits the final model. We also visualize the effectiveness of InCDP by comparing the feature distributions with and without InCDP in Fig.~\ref{fig:InCDP}. As can be seen, the feature distribution of the positive and negative samples are entangled in the model (5) without InCDP, as shown in Fig.~\ref{fig:InCDP}~(a). When equipping InCDP, the feature entangled problem is greatly relieved. The positive and negative samples in the feature space have been pushed away, leading to an obvious improvement of F1 score from 0.7418 to 0.7895.

Since MIDOG2022 and MITOSIS2014 are out-of-domain (OOD) datasets. We also investigate the impact of each novelty on OOD data. The conclusions are consistent with the previous experiment on the validation set. Every single module can improve the performance of the baseline model. The model with any two associated modules is more effective than the model with the single one. And the final model achieves the best performance. The difference is that SE module plays a more important role in the OOD data, especially for MITOSIS2014 dataset with a larger domain gap. Besides, controlling the training samples by DGSB and enhancing the feature representation by InCDP can also improve the model generalizability. The final model (8) boosts the performance by 9\% on MIDOG2022 and 28\% on MITOSIS2014. 

\begin{table}[t]
	\small
	\centering
	\scriptsize
	\caption{Effectiveness of stain enhancement (SE).}
	\begin{threeparttable}
		\begin{tabular}{ccccccc}
			\toprule
			\textbf{Datasets} &\textbf{Models} &\textbf{UnSA} &\textbf{EFDMix} &\textbf{F1} &\textbf{Recall} &\textbf{Prec} \cr
			\midrule
			\multirow{4}{*}{\makecell[c]{MITOSIS\\2014}}&(1) &- &- &0.3529 &0.3214 &0.3913 \cr
			&(2) &$\checkmark$ &- &0.4342 &0.4657 &0.4068 \cr
			&(3) &- &$\checkmark$ &0.4066 &0.4218 &0.3926 \cr
			&(4) &$\checkmark$ &$\checkmark$ &\textbf{0.5139} &\textbf{0.6011} &\textbf{0.4488} \\
			\bottomrule
			\hspace{1mm}
		\end{tabular}
	\end{threeparttable}
	\label{ablationstudy2}
\end{table}
\begin{table}[t]
	\small
	\centering
	\scriptsize
	\caption{The sensitivity of hematoxylin-based nuclei localization.}
	\label{tab:sensitivity}
	\begin{threeparttable}
		\begin{tabular}{cccccc}
			\toprule
			\textbf{Datasets} &\textbf{\makecell[c]{MIDOG\\2021\\(val)}} &\textbf{\makecell[c]{AMIDA\\2013}} &\textbf{\makecell[c]{MITOSIS\\2014}} &\textbf{\makecell[c]{TUPAC\\2016}} &\textbf{\makecell[c]{MIDOG\\2022}}\\ 
			\midrule
			Sensitivity &0.983 &0.986 &0.964 &1.0 &0.977\\
			\toprule
		\end{tabular}
	\end{threeparttable}
\end{table} 
Next, we further investigate the impact of the SE module on the generalization ability by evaluating UnSA and EFDMix separately, as shown
in Table~\ref{ablationstudy2}. The baseline model in this experiment is the model (7) in Table~\ref{ablationstudy} with DGSB and InCDP. MITOSIS2014 with the largest domain discrepancy is introduced for this experiment. For model (2), unsupervised stain augmentation introduces more information for each specific color domain, leading to a tremendous improvement of the performance by 8\% in F1 score. EFDMix enhances the cross-domain feature representation, which also brings around 5\% improvement. When associating these two techniques, the F1 score result is further boosted to 0.5139.

\subsection{Sensitivity of Hematoxylin-based Nuclei Localization}
\begin{figure}
	\includegraphics[width=.99\linewidth]{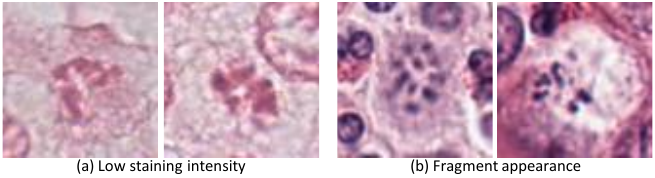}
	\caption{Two examples that are not detected by hematoxylin-based nuclei localization.}
	\label{fig:recall}
\end{figure}

One practical problem of multi-stage approaches is the sensitivity of the detection phase, which determines the upper bound of the detection performance. Thus, we discuss the pros and cons of the hematoxylin-based nuclei localization by reporting the sensitivity of each dataset in Table~\ref{tab:sensitivity}. Thanks to biological and chemical knowledge, hematoxylin-based nuclei localization maintains relatively high sensitivities in all the datasets. Fig.~\ref{fig:recall} demonstrates two examples that are not detected by our model. Fig.~\ref{fig:recall}~(a) is the image with low staining intensity, which cannot be extracted when generating the hematoxylin channel. Fig.~\ref{fig:recall}~(b) demonstrates the fragment-like appearance of mitosis which will be filtered by a threshold of nuclei area. Although it still misses a small among of mitosis, the training-free, easy-to-use, efficient and robust to color variances natures can cover up this drawback. It also filters over 99\% negative samples which greatly improves the model efficiency.

\vspace{-3mm}
\section{Conclusion}\label{sec6}
In this study, we rethink mitosis detection in data and feature aspects and propose a novel generalizable framework (MitDet) for mitosis detection. It achieves SOTA performance in one internal and four external datasets using only point annotation. Hematoxylin-based nuclei localization is proposed for training-free candidate extraction. DGSB serves for balancing the diversity, quantity and difficulty of the training samples. InCDP is proposed for diversifying the feature representation by transforming binary classes into sub-classes. To alleviate domain discrepancy, we simultaneously enhance domain-specific data and cross-domain feature distribution to improve the diversity of data and features. Considering both detection performance and annotation efficiency, MitDet is currently the best solution for generalizable mitosis detection on multi-center external data. 

Besides, MitDet also demonstrates great flexibility and extendibility. Every step can be replaced by a superior one, such as a nuclei localization approach with higher sensitivity, a smarter data sampler, or a more robust classifier. We believe that this study not only introduces a novel framework for mitosis detection but also brings new insights to this topic.

\bibliography{reference}

\begin{thebibliography}{10}
\providecommand{\url}[1]{#1}
\csname url@samestyle\endcsname
\providecommand{\newblock}{\relax}
\providecommand{\bibinfo}[2]{#2}
\providecommand{\BIBentrySTDinterwordspacing}{\spaceskip=0pt\relax}
\providecommand{\BIBentryALTinterwordstretchfactor}{4}
\providecommand{\BIBentryALTinterwordspacing}{\spaceskip=\fontdimen2\font plus
\BIBentryALTinterwordstretchfactor\fontdimen3\font minus
  \fontdimen4\font\relax}
\providecommand{\BIBforeignlanguage}[2]{{%
\expandafter\ifx\csname l@#1\endcsname\relax
\typeout{** WARNING: IEEEtran.bst: No hyphenation pattern has been}%
\typeout{** loaded for the language `#1'. Using the pattern for}%
\typeout{** the default language instead.}%
\else
\language=\csname l@#1\endcsname
\fi
#2}}
\providecommand{\BIBdecl}{\relax}
\BIBdecl

\bibitem{2010Pathological}
C.~Elston, ``Pathological prognostic factors in breast cancer. i. the value of
  histological grade in breast cancer: experience from a large study with
  long-term follow-up.'' \emph{Histopathology}, vol.~19, no.~5, pp. 403--410,
  2010.

\bibitem{tellez2018whole}
D.~Tellez, M.~Balkenhol, I.~Otte-H{\"o}ller, R.~van~de Loo, R.~Vogels, P.~Bult,
  C.~Wauters, W.~Vreuls, S.~Mol, N.~Karssemeijer \emph{et~al.}, ``Whole-slide
  mitosis detection in h\&e breast histology using phh3 as a reference to train
  distilled stain-invariant convolutional networks,'' \emph{IEEE transactions
  on medical imaging}, vol.~37, no.~9, pp. 2126--2136, 2018.

\bibitem{aubreville2023mitosis}
M.~Aubreville, N.~Stathonikos, C.~A. Bertram, R.~Klopfleisch, N.~Ter~Hoeve,
  F.~Ciompi, F.~Wilm, C.~Marzahl, T.~A. Donovan, A.~Maier \emph{et~al.},
  ``Mitosis domain generalization in histopathology images—the midog
  challenge,'' \emph{Medical Image Analysis}, vol.~84, p. 102699, 2023.

\bibitem{2008The}
A.~S. Wilkins and R.~Holliday, ``The evolution of meiosis from mitosis,''
  \emph{Genetics}, 2008.

\bibitem{Chao2019Weakly}
Chao, Xinggang, Wang, Wenyu, Liu, L.~Jan, Latecki, Junzhou, and Huang, ``Weakly
  supervised mitosis detection in breast histopathology images using concentric
  loss.'' \emph{Medical image analysis}, 2019.

\bibitem{li2022domain}
Y.~Li, Y.~Xue, L.~Li, X.~Zhang, and X.~Qian, ``Domain adaptive box-supervised
  instance segmentation network for mitosis detection,'' \emph{IEEE
  Transactions on Medical Imaging}, vol.~41, no.~9, pp. 2469--2485, 2022.

\bibitem{pan2021mitosis}
X.~Pan, Y.~Lu, R.~Lan, Z.~Liu, Z.~Qin, H.~Wang, and Z.~Liu, ``Mitosis detection
  techniques in h\&e stained breast cancer pathological images: A comprehensive
  review,'' \emph{Computers \& Electrical Engineering}, vol.~91, p. 107038,
  2021.

\bibitem{han2021contextual}
J.~Han, X.~Wang, and W.~Liu, ``Contextual prior constrained deep networks for
  mitosis detection with point annotations,'' \emph{IEEE Access}, vol.~9, pp.
  71\,954--71\,967, 2021.

\bibitem{sohail2021mitotic}
A.~Sohail, A.~Khan, H.~Nisar, S.~Tabassum, and A.~Zameer, ``Mitotic nuclei
  analysis in breast cancer histopathology images using deep ensemble
  classifier,'' \emph{Medical image analysis}, vol.~72, p. 102121, 2021.

\bibitem{mahmood2020artificial}
T.~Mahmood, M.~Arsalan, M.~Owais, M.~B. Lee, and K.~R. Park, ``Artificial
  intelligence-based mitosis detection in breast cancer histopathology images
  using faster r-cnn and deep cnns,'' \emph{Journal of clinical medicine},
  vol.~9, no.~3, p. 749, 2020.

\bibitem{ccayir2022mitnet}
S.~{\c{C}}ay{\i}r, G.~Solmaz, H.~Kusetogullari, F.~Tokat, E.~Bozaba,
  S.~Karakaya, L.~O. Iheme, E.~Tekin, {\c{C}}.~Yaz{\i}c{\i}, G.~{\"O}zsoy
  \emph{et~al.}, ``Mitnet: a novel dataset and a two-stage deep learning
  approach for mitosis recognition in whole slide images of breast cancer
  tissue,'' \emph{Neural Computing and Applications}, vol.~34, no.~20, pp.
  17\,837--17\,851, 2022.

\bibitem{wang2023generalizable}
X.~Wang, J.~Zhang, S.~Yang, J.~Xiang, F.~Luo, M.~Wang, J.~Zhang, W.~Yang,
  J.~Huang, and X.~Han, ``A generalizable and robust deep learning algorithm
  for mitosis detection in multicenter breast histopathological images,''
  \emph{Medical Image Analysis}, vol.~84, p. 102703, 2023.

\bibitem{zhao2020triple}
B.~Zhao, X.~Chen, Z.~Li, Z.~Yu, S.~Yao, L.~Yan, Y.~Wang, Z.~Liu, C.~Liang, and
  C.~Han, ``Triple u-net: Hematoxylin-aware nuclei segmentation with
  progressive dense feature aggregation,'' \emph{Medical Image Analysis},
  vol.~65, p. 101786, 2020.

\bibitem{zhao2022restainnet}
B.~Zhao, C.~Han, X.~Pan, J.~Lin, Z.~Yi, C.~Liang, X.~Chen, B.~Li, W.~Qiu, D.~Li
  \emph{et~al.}, ``Restainnet: a self-supervised digital re-stainer for stain
  normalization,'' \emph{Computers and Electrical Engineering}, vol. 103, p.
  108304, 2022.

\bibitem{zhang2022exact}
Y.~Zhang, M.~Li, R.~Li, K.~Jia, and L.~Zhang, ``Exact feature distribution
  matching for arbitrary style transfer and domain generalization,'' in
  \emph{Proceedings of the IEEE/CVF Conference on Computer Vision and Pattern
  Recognition}, 2022, pp. 8035--8045.

\bibitem{wang2022novel}
H.~Wang, Z.~Liu, R.~Lan, Z.~Liu, X.~Luo, X.~Pan, and B.~Li, ``A novel dataset
  and a deep learning method for mitosis nuclei segmentation and
  classification,'' \emph{arXiv preprint arXiv:2212.13401}, 2022.

\bibitem{li2019weakly}
C.~Li, X.~Wang, W.~Liu, L.~J. Latecki, B.~Wang, and J.~Huang, ``Weakly
  supervised mitosis detection in breast histopathology images using concentric
  loss,'' \emph{Medical image analysis}, vol.~53, pp. 165--178, 2019.

\bibitem{long2015fully}
J.~Long, E.~Shelhamer, and T.~Darrell, ``Fully convolutional networks for
  semantic segmentation,'' in \emph{Proceedings of the IEEE conference on
  computer vision and pattern recognition}, 2015, pp. 3431--3440.

\bibitem{yang2020fda}
Y.~Yang and S.~Soatto, ``Fda: Fourier domain adaptation for semantic
  segmentation,'' in \emph{Proceedings of the IEEE/CVF Conference on Computer
  Vision and Pattern Recognition}, 2020, pp. 4085--4095.

\bibitem{graham2019hover}
S.~Graham, Q.~D. Vu, S.~E.~A. Raza, A.~Azam, Y.~W. Tsang, J.~T. Kwak, and
  N.~Rajpoot, ``Hover-net: Simultaneous segmentation and classification of
  nuclei in multi-tissue histology images,'' \emph{Medical Image Analysis},
  vol.~58, p. 101563, 2019.

\bibitem{ren2015faster}
S.~Ren, K.~He, R.~Girshick, and J.~Sun, ``Faster r-cnn: Towards real-time
  object detection with region proposal networks,'' \emph{Advances in neural
  information processing systems}, vol.~28, 2015.

\bibitem{he2017mask}
K.~He, G.~Gkioxari, P.~Doll{\'a}r, and R.~Girshick, ``Mask r-cnn,'' in
  \emph{Proceedings of the IEEE international conference on computer vision},
  2017, pp. 2961--2969.

\bibitem{zhou2022domain}
K.~Zhou, Z.~Liu, Y.~Qiao, T.~Xiang, and C.~C. Loy, ``Domain generalization: A
  survey,'' \emph{IEEE Transactions on Pattern Analysis and Machine
  Intelligence}, 2022.

\bibitem{li2020domain}
H.~Li, Y.~Wang, R.~Wan, S.~Wang, T.-Q. Li, and A.~Kot, ``Domain generalization
  for medical imaging classification with linear-dependency regularization,''
  \emph{Advances in Neural Information Processing Systems}, vol.~33, pp.
  3118--3129, 2020.

\bibitem{zhang2020generalizing}
L.~Zhang, X.~Wang, D.~Yang, T.~Sanford, S.~Harmon, B.~Turkbey, B.~J. Wood,
  H.~Roth, A.~Myronenko, D.~Xu \emph{et~al.}, ``Generalizing deep learning for
  medical image segmentation to unseen domains via deep stacked
  transformation,'' \emph{IEEE transactions on medical imaging}, vol.~39,
  no.~7, pp. 2531--2540, 2020.

\bibitem{li2022domaingeneralization}
C.~Li, X.~Lin, Y.~Mao, W.~Lin, Q.~Qi, X.~Ding, Y.~Huang, D.~Liang, and Y.~Yu,
  ``Domain generalization on medical imaging classification using episodic
  training with task augmentation,'' \emph{Computers in biology and medicine},
  vol. 141, p. 105144, 2022.

\bibitem{HAN2022102481}
C.~Han, H.~Yao, B.~Zhao, Z.~Li, Z.~Shi, L.~Wu, X.~Chen, J.~Qu, K.~Zhao, R.~Lan,
  C.~Liang, X.~Pan, and Z.~Liu, ``Meta multi-task nuclei segmentation with
  fewer training samples,'' \emph{Medical Image Analysis}, p. 102481, 2022.

\bibitem{zhao2021robust}
X.~Zhao, A.~Sicilia, D.~S. Minhas, E.~E. O’Connor, H.~J. Aizenstein, W.~E.
  Klunk, D.~L. Tudorascu, and S.~J. Hwang, ``Robust white matter hyperintensity
  segmentation on unseen domain,'' in \emph{2021 IEEE 18th International
  Symposium on Biomedical Imaging (ISBI)}.\hskip 1em plus 0.5em minus
  0.4em\relax IEEE, 2021, pp. 1047--1051.

\bibitem{ruifrok2001quantification}
A.~C. Ruifrok, D.~A. Johnston \emph{et~al.}, ``Quantification of histochemical
  staining by color deconvolution,'' \emph{Analytical and quantitative cytology
  and histology}, vol.~23, no.~4, pp. 291--299, 2001.

\bibitem{wu2019wider}
Z.~Wu, C.~Shen, and A.~Van Den~Hengel, ``Wider or deeper: Revisiting the resnet
  model for visual recognition,'' \emph{Pattern Recognition}, vol.~90, pp.
  119--133, 2019.

\bibitem{chen2021exploring}
X.~Chen and K.~He, ``Exploring simple siamese representation learning,'' in
  \emph{Proceedings of the IEEE/CVF conference on computer vision and pattern
  recognition}, 2021, pp. 15\,750--15\,758.

\bibitem{veta2015assessment}
M.~Veta, P.~J. Van~Diest, S.~M. Willems, H.~Wang, A.~Madabhushi, A.~Cruz-Roa,
  F.~Gonzalez, A.~B. Larsen, J.~S. Vestergaard, A.~B. Dahl \emph{et~al.},
  ``Assessment of algorithms for mitosis detection in breast cancer
  histopathology images,'' \emph{Medical image analysis}, vol.~20, no.~1, pp.
  237--248, 2015.

\bibitem{li2018deepmitosis}
C.~Li, X.~Wang, W.~Liu, and L.~J. Latecki, ``Deepmitosis: Mitosis detection via
  deep detection, verification and segmentation networks,'' \emph{Medical image
  analysis}, vol.~45, pp. 121--133, 2018.

\bibitem{sebai2020maskmitosis}
M.~Sebai, X.~Wang, and T.~Wang, ``Maskmitosis: a deep learning framework for
  fully supervised, weakly supervised, and unsupervised mitosis detection in
  histopathology images,'' \emph{Medical \& Biological Engineering \&
  Computing}, vol.~58, pp. 1603--1623, 2020.

\bibitem{aubreville2021quantifying}
M.~Aubreville, C.~Bertram, M.~Veta, R.~Klopfleisch, N.~Stathonikos,
  K.~Breininger, N.~ter Hoeve, F.~Ciompi, and A.~Maier, ``Quantifying the
  scanner-induced domain gap in mitosis detection,'' \emph{arXiv preprint
  arXiv:2103.16515}, 2021.

\bibitem{wollmann2017deep}
T.~Wollmann and K.~Rohr, ``Deep residual hough voting for mitotic cell
  detection in histopathology images,'' in \emph{2017 IEEE 14th International
  Symposium on Biomedical Imaging (ISBI 2017)}.\hskip 1em plus 0.5em minus
  0.4em\relax IEEE, 2017, pp. 341--344.

\bibitem{lei2020attention}
H.~Lei, S.~Liu, A.~Elazab, X.~Gong, and B.~Lei, ``Attention-guided multi-branch
  convolutional neural network for mitosis detection from histopathological
  images,'' \emph{IEEE Journal of Biomedical and Health Informatics}, vol.~25,
  no.~2, pp. 358--370, 2020.

\bibitem{lafarge2017domain}
M.~W. Lafarge, J.~P. Pluim, K.~A. Eppenhof, P.~Moeskops, and M.~Veta,
  ``Domain-adversarial neural networks to address the appearance variability of
  histopathology images,'' in \emph{Deep Learning in Medical Image Analysis and
  Multimodal Learning for Clinical Decision Support: Third International
  Workshop, DLMIA 2017, and 7th International Workshop}.\hskip 1em plus 0.5em
  minus 0.4em\relax Springer, 2017, pp. 83--91.

\bibitem{akram2018leveraging}
S.~U. Akram, T.~Qaiser, S.~Graham, J.~Kannala, J.~Heikkil{\"a}, and N.~Rajpoot,
  ``Leveraging unlabeled whole-slide-images for mitosis detection,'' in
  \emph{Computational Pathology and Ophthalmic Medical Image Analysis: First
  International Workshop, COMPAY 2018, and 5th International Workshop}.\hskip
  1em plus 0.5em minus 0.4em\relax Springer, 2018, pp. 69--77.

\end{thebibliography}
\bibliographystyle{IEEEtran}

\end{document}